\documentclass[sigconf,natbib=true]{acmart}
 
\usepackage{setspace}
\usepackage{amsmath}               
  {
      \theoremstyle{plain}
      \newtheorem{assumption}{Assumption}
  }
\usepackage{enumitem}
\usepackage{tikz}
\usepackage{subcaption}
\usepackage{pgfplots}
\usetikzlibrary{arrows,positioning,automata,calc,shapes}
\pgfplotsset{compat=newest, scaled z ticks=false} 
\pgfplotsset{plot coordinates/math parser=false}
\newlength\figureheight 
\newlength\figurewidth
\usepackage{tikz}
\usepackage{caption}
\usepackage{filecontents}
\usepackage{url}
\usepackage{amsmath}
\usepackage{graphicx}
\usepackage{url}

\usepackage{amsfonts,amssymb}
\usepackage{color, soul}
\usepackage{mathrsfs}
\usepackage{cleveref}
\usepackage{multirow}
\usepackage{wrapfig}
\usepackage{amsthm}
\usepackage{algorithm}
\usepackage{algorithmic}
\usepackage{bm}
\usepackage{mdframed}
\usepackage{tikz}

\newcommand{\indep}{\perp \!\!\! \perp}
\AtBeginDocument{%
  \providecommand\BibTeX{{%
    \normalfont B\kern-0.5em{\scshape i\kern-0.25em b}\kern-0.8em\TeX}}}

\copyrightyear{2024}
\acmYear{2024}
\setcopyright{rightsretained}
\acmConference[KDD '24]{Proceedings of the 30th ACM SIGKDD Conference on Knowledge Discovery and Data Mining}{August 25--29, 2024}{Barcelona, Spain}
\acmBooktitle{Proceedings of the 30th ACM SIGKDD Conference on Knowledge Discovery and Data Mining (KDD '24), August 25--29, 2024, Barcelona, Spain}\acmDOI{10.1145/3637528.3671450.}
\acmISBN{979-8-4007-0490-1/24/08}

\begin{document}

\author{Yaochen Zhu}
\affiliation{%
\institution{University of Virginia}
\city{}
\state{}
\country{}
}
\email{uqp4qh@virginia.edu}

\author{Yinhan He}
\affiliation{%
\institution{University of Virginia}
\city{}
\state{}
\country{}
}
\email{nee7ne@virginia.edu}

\author{Jing Ma}
\affiliation{%
\institution{Case Western Reserve University}
\city{}
\state{}
\country{}
}
\email{jing.ma@case.edu}

\author{Mengxuan Hu}
\affiliation{%
\institution{University of Virginia}
\city{}
\state{}
\country{}
}
\email{qtq7su@virginia.edu}

\author{Sheng Li}
\affiliation{%
\institution{University of Virginia}
\city{}
\state{}
\country{}
}
\email{shengli@virginia.edu}

\author{Jundong Li}
\affiliation{%
\institution{University of Virginia}
\city{}
\state{}
\country{}
}
\email{jundong@virginia.edu}

\fancyhead{}

\title{Causal Inference with Latent Variables: Recent Advances \\and Future Prospectives}

\begin{abstract}

\noindent Causality lays the foundation for the trajectory of our world. Causal inference (CI), which aims to infer intrinsic causal relations among variables of interest, has emerged as a crucial research topic. Nevertheless, the lack of observation of important variables (e.g., confounders, mediators, exogenous variables, etc.) severely compromises the reliability of CI methods. The issue may arise from the inherent difficulty in measuring the variables. Additionally, in observational studies where variables are passively recorded, certain covariates might be inadvertently omitted by the experimenter. Depending on the type of unobserved variables and the specific CI task, various consequences can be incurred if these latent variables are carelessly handled, such as biased estimation of causal effects, incomplete understanding of causal mechanisms, lack of individual-level causal consideration, etc. In this survey, we provide a comprehensive review of recent developments in CI with latent variables. We start by discussing traditional CI techniques when variables of interest are assumed to be fully observed. Afterward, under the taxonomy of circumvention and inference-based methods, we provide an in-depth discussion of various CI strategies to handle latent variables, covering the tasks of causal effect estimation, mediation analysis, counterfactual reasoning, and causal discovery. Furthermore, we generalize the discussion to graph data where interference among units may exist. Finally, we offer fresh aspects for further advancement of CI with latent variables, especially new opportunities in the era of large language models (LLMs).

\end{abstract}



\maketitle

\section{Introduction}

\noindent Our world is a woven web of causes and effects, where everything that occurs is the consequence of some prior actions \cite{anscombe2018causality,zhu2023causal}. For example, my headache disappeared because of the aspirin I took this afternoon, and I gained muscle because I worked out regularly every day. From the levity of language, we may be under the illusion that reasoning with causality from experiences can be simple and straightforward. However, formal causal inference did not emerge until decades ago, which enabled rigorously derivation of causal relationships of interest from the observational data \cite{rubin2005causal}.

In hindsight, what prevented the emergence of formal causal inference (CI) is the lack of mathematical language to describe causality \cite{pearl2018book}. One tempting choice is to use conditional distributions from probability theory for causal reasoning \cite{breiman1992probability}. For example, if an event $T$ causes another event $Y$ (where $T,Y=1$ means that the event \textit{happened} and 0 otherwise), we usually have $p(Y = 1|T =1) > p(Y =1 | T =0)$. However, if we use the converse, i.e., the increase of probability, to denote causality, \textit{correlation} can be easily mistaken for \textit{causation}. For example, we can observe that people eat more ice cream when they wear fewer clothes. However, the former is clearly not a cause for the latter, as both are caused by a third variable: hot weather. Here, the issue lies in the fact that $T=1$ in the conditional distribution means that $T$ is passively observed, but what makes the relation between $T$ and $Y$ causal is that $Y$ will happen if we \textit{make} $T$ happen. That is why Rubin claimed that "there is no causality without intervention" and introduced the potential outcome $Y(T=1)$ to describe the event $Y$ if $T=1$ is \textit{made} to happen for all population \cite{imbens2015causal}. Similarly, Pearl introduced the \textit{do}-operator, where $p(Y|do(T))$ denotes the distribution of $Y$ if we make the event $T$ happen instead of observing it passively \cite{pearl2009causality}. 

With new symbols defined to facilitate causal reasoning, various causal questions can be formed in a rigorous manner\footnote{Here, we only use ATE as an example, where the formulations of other CI tasks with potential outcome or do-operator are detailed in the subsequent sections.}. One common CI task is average treatment effect (ATE) estimation \cite{angrist1995identification}, which aims to estimate the expected influence of an event ($T$) on another ($Y$), e.g., the change of recovery rate $Y$ if drug $T$ is prescribed to all patients. Since ATE compares the outcomes of two interventions, i.e., treatment/no treatment, it can be directly formulated as $\mathbb{E}[Y(T=1)] - \mathbb{E}[Y(T=0)]$ (via potential outcome) or $\mathbb{E}[Y|do(T=1)] - \mathbb{E}[Y|do(T=0)]$ (via \textit{do}-operator). In addition, causal mediation analysis \cite{imai2010general} is also feasible via the new symbols, which aims to determine the fine-grained causal effect of $T$ on $Y$ mediated by other factors. For example, if we know that drug $T$ cures the disease by reducing the blood pressure $Z$ but it also thickens the blood vessel wall, we can define the causal effects of $T$ on $Y$ mediated by $Z$ as the effect as if the drug has no side effect. Furthermore, the individual level causal effect also becomes tractable \cite{kusner2017counterfactual}. For example, for Alice, who has received the treatment and survived, we can formulate the question "would she also survive if no treatment had been provided?". Finally, we can even formulate causal discovery with the new symbols \cite{spirtes2016causal}, where causal relations among variables of interest (e.g., treatment, mediators, individual factors) can be automatically discovered from data.

Nevertheless, representing causal questions with new symbols is not enough. After all, directly obtaining the causal estimand $Y(T)$ or $p(Y|do(T))$ requires intervention upon $T$, which is not always feasible. One strategy is to simulate interventions with randomized experiment (RE) \cite{greenland1990randomization}, where the randomization ensures that treatment $T$ is the only contributor to the variation of $Y$. However, RE can be expensive and unethical (e.g., we cannot randomly decide whether or not to give drugs to patients). Therefore, CI with observational studies gains more attention, where the experimenter has no manipulation over the treatment assignment. The aim is to show that if certain assumptions hold for the data (i.e., identification criteria), causal estimand with causal symbols can still be calculated with conditional relations measurable in the collected data. For example, if all confounders (i.e., factors that simultaneously affect cause $T$ and effect $Y$) are observed and recorded, backdoor adjustment \cite{tian2002general} and propensity score weighting \cite{rosenbaum1983central} can be used to estimate ATE. In addition, if the mediator of interest $M$ is measured, the path-specific effect of $T$ on $Y$ mediated by $M$ can be obtained under certain identification criteria \cite{imai2010identification}. If exogenous variables (e.g., individual factors not considered as the main variables of interest) are known, individual-level counterfactuals can be calculated \cite{pearl2009causality}. Finally, if all variables of interest are known, mature algorithms such as the PC algorithm \cite{spirtes2000causation} are off-the-shelf for causal discovery.

However, important variables for CI can be latent, which hinders the reliability of existing CI techniques \cite{spirtes2013causal}. The issue lies primarily in two folds: \textbf{\textit{(i)}} First, certain variables can be intrinsically difficult to measure, e.g., the socioeconomic status of a patient, which is a crucial confounder for drug effect evaluation \cite{louizos2017causal}. \textbf{\textit{(ii)}} In addition, in the observational study, important covariates for CI may not be recorded in the collected data \cite{nichols2007causal}. The consequences of carelessly handling latent variables for CI can be multi-faceted. First, unobserved confounders can lead to bias in ATE estimation \cite{chu2023causal}; for example, if the severity of disease is not considered, we may erroneously conclude that an effective drug lowers the recovery rate, as more severe patients tend to be treated with the drug. In addition, missing important mediators could result in an incomplete understanding of the causal mechanism \cite{muthen2015causal}. For example, the debate over the causal relation between tobacco smoking and lung cancer was not resolved until the mediator Tar deposit was determined to cause lung cancer for smokers \cite{sasco2004tobacco}. Exogenous variables are usually considered as noise and are not explicitly included in the observational data \cite{kusner2017counterfactual}. However, without them, individual differences in treatment effects cannot be estimated, which hinders personalized counterfactual analysis. Finally, if not all variables of interest are available, causal discovery would be impossible \cite{cai2019causal}. 

Recent years have witnessed a plethora of works on causal inference with latent variables \cite{louizos2017causal}. Generally, the methods can be categorized into two classes: \textbf{\textit{(i)}} \textit{Circumvention-based Methods} and \textbf{\textit{(ii)}} \textit{Inference-based Methods}. Circumvention-based strategies eschew direct modeling of latent variables; instead, they show that under certain stringent assumptions/conditions,  latent variables can be avoided while the causal estimand can still be identified with observational data. However, there is no free lunch, and the price being paid could be the requirement to measure more variables (where errors could be introduced) \cite{fulcher2020robust} and an increase in estimation variance \cite{baiocchi2014instrumental}. Inference-based methods, in contrast, explicitly model the latent variables based on the observations. This usually includes proxy of the latent variables (e.g., their noisy observations). However, latent variables may not be identifiable given the observed data, where bias can still remain in the causal estimations \cite{khemakhem2020variational}. In addition, the proxy of latent variables may contain undesirable components, and carelessly ignoring them can ruin the estimation results \cite{montgomery2018conditioning}. Both strategies on the main CI tasks, as well as their generalization to graph data where interference exists, will be thoroughly discussed in this survey. The distinctive contribution of us can be concretely summarized into three folds as follows:
\begin{itemize}[leftmargin=0.5cm]
    \item \textit{\textbf{Timely Topic.}} CI with latent variable is an important topic while scattered in different CI areas. This survey provides a timely and comprehensive review of the state-of-the-art.
    \item \textit{\textbf{Novel Taxonomy.}} We provide novel taxonomy on existing CI methods to address latent variables, where two main categories of methods on four CI tasks are thoroughly discussed.
    \item \textit{\textbf{New Hope.}} Based on existing techniques, we provide insights into the future advancement of CI with latent variables, especially the new opportunities with large language models (LLM).
\end{itemize}

\section{Preliminaries}
\label{sec:prelim}

\subsection{Symbol System}

For most CI tasks, there are two main variables of interest, i.e., treatment $T$ and outcome $Y$, on which the causal relation is scrutinized.  We consider $T$ as a binary variable by default, but the cases of continuous/multiple/high-dimensional treatments will also be covered in detail. The outcome $Y$ can be arbitrary results of interest under the potential causal influence of $T$. In addition, we use $X$ to denote other observed covariates in the system, which may have certain causal relations with $T$ and $Y$ depending on the context. 

\subsection{Rubin's Causal Model}

To study the causal relation between treatment $T$ and outcome $Y$, Rubin's causal model (SCM) starts by comparing individual-level counterfactuals, i.e., for unit $i$, what the outcome $Y$ is if the unit is treated ($T=1$) or is not treated ($T=0$). Although the two results cannot be observed for the same unit $i$ simultaneously, we can still hypothetically define them as potential outcomes as follows:
\begin{definition}
\textbf{\textit{(Potential Outcome)}.} We use the notations $\{Y_{i}(T=1)$, $Y_{i}(T=0)\}$ (which are shortened as $Y_{i}(1)$, $Y_{i}(0)$ if the treatment is clear from the context) to denote the potential outcomes (PO) of $Y$ for unit $i$ if the treatment $T=1$ or $0$ is imposed on the unit.
\end{definition}
Accordingly, \textbf{R.V.} $Y(T=1)$, $Y(T=0)$ reason with the distribution of the POs if all units are uniformly treated or non-treated (i.e., interventions). However, $Y(T=1)$, $Y(T=0)$ cannot be obtained due to lack of individual counterfactuals. To estimate $Y(T=1)$, $Y(T=0)$, most strategies need to collect the outcomes of two groups of treated and non-treated units. Here, we use conditional R.V. $Y|T=1$ and $Y|T=0$ to denote the distribution of $Y$ for the two groups. Only in rare cases, e.g., randomized experiments, can $Y|T=t$ provide an unbiased estimate for $Y(T=t)$. In other cases, the purpose of RCM is to show that under certain assumptions, causal estimand with PO can be reduced to conditional relations measurable in the data (usually involving other covariates $X$).

\subsection{Structural Causal Model}
\label{sec:scm}
Pearl's structural causal model (SCM), in contrast, reasons with causality via a pre-defined direct acyclic causal graph $\mathbf{G}$ that encodes the belief of causal relations among variables of interest \cite{pearl2009causality}. Based on the causal graph, SCM can be formally defined as follows:
\begin{definition}
\textbf{\textit{(SCM)}.}  Structural causal model (SCM) can be defined as a triplet of sets $(\mathcal{U}, \mathcal{V}, \mathcal{F})$, where $\mathcal{U}$ is the set of latent exogenous variables, $\mathcal{V}$ is a set of observed endogenous variables, and $\mathcal{F}$ is a set of structural equations. For an endogenous variable $V \in \mathcal{V}$, we have $V = f_{V}(\mathcal{A}_{U}(V), \mathcal{A}_{V}(V))$, where $\mathcal{A}_{U}(V), \mathcal{A}_{V}(V)$ are the exogenous, endogenous parents of $V$ in $\mathbf{G}$, respectively. 
\end{definition}

\begin{figure}[t]
\centering 
\includegraphics[width=0.45\textwidth]{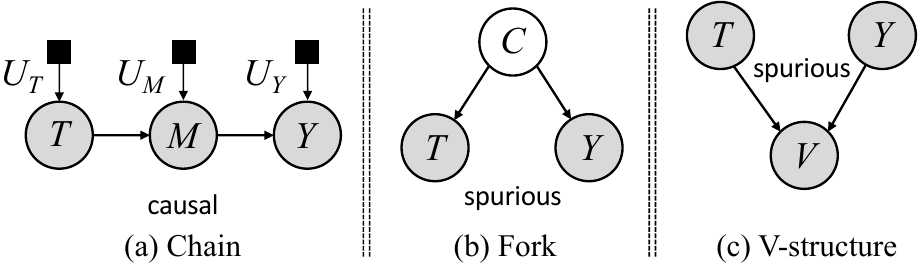}
\vspace{-2mm}
\caption{Atomic structures of SCM, where mutually independent exogenous variables $\mathcal{U}$ are omitted for (b) and (c).}
\vspace{-3mm}
\label{fig:scm}
\end{figure}

\noindent In SCM, each unit $i$ is associated with a set of exogenous variables $\mathcal{U} = \mathcal{U}_{i}$ that causally determines the endogenous variables, e.g., $T$, $Y$, $X$. The prior for $\mathcal{U}$ is $p(\mathcal{U})$. Mutually-independent exogenous variables are usually ignored when average causal effects are considered, but they are vital for counterfactual reasoning since they represent unit variations. Three atomic structures exist in a causal graph (Fig. \ref{fig:scm}): \textbf{\textit{(i)}} chains $T \rightarrow M \rightarrow Y$, \textbf{\textit{(ii)}} forks $T \leftarrow C \rightarrow Y$, and \textbf{\textit{(iii)}} V-structure $T \rightarrow V \leftarrow Y$. $T$ and $Y$ are correlated if \textit{mediator} $M$ is not unobserved for chains (causal), \textit{confounder} $C$ is not observed for forks (not causal), and \textit{collider} $V$ is observed for V-structures (not causal). Therefore, to distinguish causation from correlation, Pearl introduces the \textit{do}-operator, where $p(Y|do(T=t))$ means that we set $T=t$ as an intervention and calculate $Y$ via $f_{Y}(T=t, \mathcal{A}_{U}(Y), \mathcal{A}_{V}(Y) / T)$, regardless of observed parents of $T$.

\subsection{Connections between SCM and RCM}
\label{sec:connect}

If an SCM is correctly specified, potential outcome $Y_{i}(T=t)$ can be derived by \textbf{\textit{(i)}} replacing the structural equation $f_{T}$ in $\mathcal{F}$ with $f^{do}_{T} = t$ (i.e., intervention), which results in a new set of structural equations $\mathcal{F}^{do}$ , \textbf{\textit{(ii)}} setting the exogenous variables $\mathcal{U}=\mathcal{U}_{i}$ (i.e., the individual factors for unit $i$), and \textbf{\textit{(iii)}} calculating the outcome $Y$ based on $\mathcal{U}_{i}$ and the new structural equations $\mathcal{F}^{do}$ . R.V. $Y(T=t)$ can be similarly derived by using the prior of $\mathcal{U}$, i.e., $p(\mathcal{U})$, instead of $\mathcal{U}_{i}$. Therefore, the two frameworks are fundamentally equivalent. 

\subsection{Overview of Causal Inference Tasks}

\subsubsection{\textbf{Treatment Effect Estimation}} Treatment effect estimation aims to quantitatively measure the causal influence of treatment $T$ (e.g., drug) on outcome $Y$ (e.g., survival rate). The most commonly used metric is the \textbf{average treatment effect (ATE)} \cite{angrist1995identification}, which is the expected causal effect of $T$ on $Y$ for the entire population. ATE can be formulated via the two frameworks as follows:
\begin{equation}
\label{eq:ate}
ATE = \mathbb{E}[Y|do(T=1)] - \mathbb{E}[Y|do(T=0)] = \mathbb{E}[Y(T=1) - Y(T=0)].
\end{equation}
For a pretreatment variable $X$ (e.g., age), we can also define the \textbf{conditional average treatment effect (CATE)} \cite{abrevaya2015estimating} as follows:
\begin{equation}
CATE(X) = \mathbb{E}[Y|do(T=1),X] - \mathbb{E}[Y|do(T=0),X],
\end{equation}
which is important when the treatment effect is heterogeneous, i.e., different sub-populations $X$ have different responses to treatment.

\subsubsection{\textbf{Causal Mediation Analysis}} Causal mediation analysis aims to quantitatively study the fine-grained causal relationship between treatment $T$ (e.g., drug) and outcome $Y$ (e.g., survival rate) mediated by certain factors $M$ (e.g., blood pressure) \cite{imai2010general}. When there is only one mediator, the most common metric is the \textbf{natural indirect effect (NIE)}, which can be formulated as follows \cite{pearl2022direct}:
\begin{equation}
\label{eq:nie}
NIE = \mathbb{E}[Y(M(T=1), T=0))] - \mathbb{E}[Y(M(T=0), T=0)].
\end{equation}
Here, $Y(M(T=t), T=0))$ is a nested potential outcome (NPO) denoting three interventions: \textbf{\textit{(i)}} $T \leftarrow t$ along the path $M \leftarrow T$, \textbf{\textit{(ii)}} $T \leftarrow 0$ along the path $Y \leftarrow T$, and \textbf{\textit{(iii)}} $M \leftarrow M(T=t)$ along path $Y \leftarrow M$. NIE excludes the direct effect of $T$ along path $T \rightarrow Y$, while enabling the indirect effect of $T$ mediated by $M$. Furthermore, NIE can be generalized to an arbitrary causal path, i.e., path-specific causal effect, which can be defined in a similar way via NPO \cite{imai2010general}.

\subsubsection{\textbf{Counterfactual Reasoning}} Counterfactuals can be broadly defined as causal estimands (represented by $do$-operator or potential outcomes) that \textit{contradict} the factual observations (represented by conditional distributions). For example, the  \textbf{average treatment effect on the treated (ATT)} is defined as follows:
\begin{equation}
\label{eq:count}
ATT = \mathbb{E}[Y(1)|T=1] - \mathbb{E}[Y(0)|T=1]\footnote{Here, please note \textit{\textbf{consistency}} is always assumed, i.e., $\mathbb{E}[Y(1)|T=1] = \mathbb{E}[Y|T=1]$.}.
\end{equation}
Here, $\mathbb{E}[Y(0)|T=1]$ in Eq. (\ref{eq:count}) denotes the expected outcome $Y$ in a counterfactual world where the treated units (denoted by the condition $T=1$) had not been treated (denoted by $Y(T=0)$).

\subsubsection{\textbf{Causal Discovery}}

Given variables of interest, causal discovery aims to recover the causal graph $\mathbf{G}$ given the observed data, such that the parent nodes are the direct cause of the child \cite{spirtes2016causal}. 

\section{Treatment Effect Estimation}
\label{sec:conf}

\noindent  In this section, we discuss the treatment effect estimation with latent variables. Specifically, we mainly focus on the latent confounders, which can systematically bias the estimation if handled carelessly. We first introduce traditional methods where confounders are assumed to be fully observed. We then discuss the \textit{circumvention}-based and \textit{inference}-based strategies to handle latent confounders.

\subsection{Brief Review of Traditional Methods}
\label{sec:conf_trad}
Traditional treatment effect estimation methods assume away the latent confounders via the following ignorability assumption:
\begin{assumption}
\label{ass:si}
\textbf{(Ignorability).}  $Y(T=0,1) \indep T | X.$  
\end{assumption}
\noindent Combined with other common assumptions for CI (e.g., positivity, non-interference, etc. see \cite{imbens2015causal}), ATE and CATE can be identified from observational data by controlling $X$ as follows:
\begin{equation}
\label{eq:back}
CATE(X) = \mathbb{E}[Y|T=1,X] - \mathbb{E}[Y|T=0,X], ATE = \mathbb{E}_{p(X)}[CATE(X)]
\end{equation}
From the SCM's perspective, $X$ blocks all backdoor paths that lead to spurious correlations between $T$ and $Y$ (see Section \ref{sec:scm}), such that in each stratum of $X=x$, the correlation between $T$ and $Y$ is causal. Based on Eq. (\ref{eq:back}), adjustment-based methods use non-parametric methods \cite{athey2016recursive} or fit parametric models $f(t, X)$ (which will be denoted as $f_{t}(X)$ if different models are used for different $T$) to estimate $\mathbb{E}[Y|T=t, X]$, including linear models \cite{imbens2015causal}, tree-based methods \cite{wager2018estimation}, and deep neural networks (DNN) \cite{shalit2017estimating}. Another line of methods reweights samples via inverse propensity score $\mathbb{E}[T=t|X]$, such that they can be viewed as pseudo-random samples.

\subsection{Circumvention-based Methods}

However, if important confounders $C$ are missing from the observed covariates $X$, Assumption \ref{ass:si} failed, and Eq. (\ref{eq:back}) is a biased estimation for C/ATE. To address the latent confounding bias, circumvention-based methods show that, under certain stringent conditions, causal effects can still be unbiasedly estimated without the direct or indirect measurement of latent confounders or their proxies. 

\subsubsection{\textbf{Small Randomized Data}}
\label{sec:fcb_method}
If a small amount of randomized data is available (which cannot be directly used to estimate CATE due to high variance), we can use them to correct the bias in large-scale observational data with latent confounders \cite{taddy2016nonparametric, peysakhovich2016combining,van2024adaptive,colnet2024causal}. One exemplar work is \cite{kallus2018removing}, which first fits a biased CATE estimator $f^{obs}_{t}(X)$ on the observational data as Eq. (\ref{eq:back}) and correct the bias with another estimator $e^{exp}_{t}(X)$ fitting on the error of $f^{obs}_{t}(X)$ evaluated on the randomized data. Since the value of the bias is usually smaller than the CATE, the estimation variance can be reduced compared to directly fitting the CATE estimator on the small-scale randomized data. In contrast, \citet{yang2020improved} directly tackles confounding bias in the observational data. They define the latent confounding bias with the confounding function as follows: 
\begin{equation}
\lambda(X) = \mathbb{E}^{obs}[Y(0)|T=1, X]-\mathbb{E}^{obs}[Y(0)|T=0, X],
\end{equation}
which measures the systematic difference of the expected baseline PO $Y(0)$ between the treatment/non-treatment group in the observational data. They further show that under the \textit{\textbf{transportability}} assumption, i.e., the treatment effect is the same between the randomized samples and the treated samples in the observational data, the confounding bias $\lambda(X)$ can be estimated as follows\footnote{Proof is straightforward with consistency and transportability assumptions.}:
\begin{equation}
\begin{aligned}
\lambda(X) = &(\mathbb{E}^{obs}[Y|T=1, X]-\mathbb{E}^{obs}[Y|T=0, X]) -\\
&(\mathbb{E}^{exp}[Y|T=1, X]-\mathbb{E}^{exp}[Y|T=0, X]).
\end{aligned}
\end{equation}
Correction of the above bias upon the biased CATE estimator fit on the observational data leads to an unbiased CATE estimator with variance lower than direct estimation with randomized data. Recently, \citet{wu2022integrative} improved over \cite{yang2020improved} by adopting the R-learner \cite{nie2021quasi} to model the confounding function, which allows flexible ML models such as trees and DNNs as the estimator.

The advantage of using randomized data to tackle latent confounding is that the randomized data are guaranteed to be unbiased (but with high variance due to small scale). However, these methods fail in the case where even a small number of randomized samples cannot be obtained, e.g., when the dataset was collected in the past.

\begin{figure}[t]
\centering 
\includegraphics[width=0.35\textwidth]{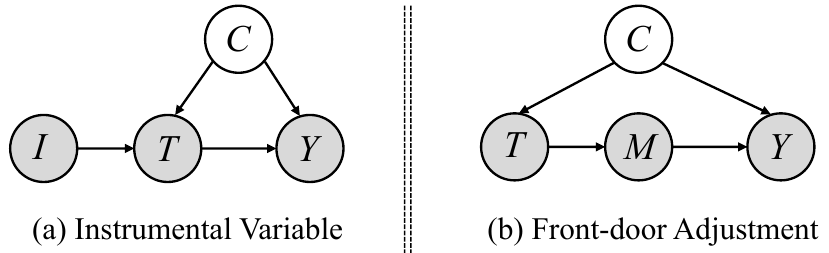}
\vspace{-2mm}
\caption{SCM for IV methods and front-door adjustment}
\vspace{-4mm}
\label{fig:ate12}
\end{figure}
\subsubsection{\textbf{Instrumental Variable}}
\label{sec:iv}
If randomized data are not possible, we can use instrumental variables (IV) to "extract" pseudo-randomized data embedded inside the observational dataset to unbiasedly estimate ATE/CATE. Formally, IV is defined as follows:
\begin{definition}
\label{def_iv}
\textbf{\textit{(Instrumental Variable, IV)}} A variable that \textbf{\textit{(i)}} has no confounding with the outcome $Y$, \textbf{\textit{(ii)}} affects the treatment $T$ (relevance), \textbf{\textit{(iii)}} affects the outcome $Y$ only through $T$ (restriction).
\end{definition}
\noindent 
For a binary IV, ATE can be unbiasedly estimated via \cite{greenland2000introduction}:
\begin{equation}
\hat{ATE} = (\mathbb{E}[Y|I=1]-\mathbb{E}[Y|I=0]) / (\mathbb{E}[T|I=1]-\mathbb{E}[T|I=0]),
\end{equation}
if we view $I$ as the assigned treatment and $T$ as the treatment received, the numerator can be viewed as the intention-to-treat effect of the treatment assignment ($I$) on outcome ($Y$), and the denominator as the compliance with the assigned treatment. General IV-based methods follow a similar two-stage procedure. Assuming linear causal relations, the two-stage least squares algorithm (2SLS) \textbf{\textit{(i)}} first calculates the conditional mean of the treatment $T$ given the IV $I$, i.e., $\hat{T} = \mathbb{E}[T|I]$, and \textbf{\textit{(ii)}} regresses $Y$ on $\hat{T}$, where the coefficient gives the causal relation between $Y$ and $T$ \cite{angrist1996identification}. Afterward, efforts have been devoted to generalizing 2SLS to nonlinear cases \cite{singh2019kernel,dikkala2020minimax,muandet2020dual}. For example, Deep IV \cite{hartford2017deep} estimates the conditional density in $\hat{T} = \mathbb{E}[T|I, X]$ from stage \textbf{\textit{(i)}} with categorical distribution (for discrete $T$) or mixture of Gaussian distribution (for continuous $T$) parameterized by DNN, and predicts the outcome $Y$ in stage \textbf{\textit{(ii)}} via another DNN $\hat{Y} = f_{nn}(\hat{T}, X)$. However, to make the objective optimizable, Deep IV assumes simple distributions, which fail when the treatment $T$ is high dimensional. To address this issue, \citet{bennett2019deep} proposed to use a generalized method of moments to allow more flexible DNNs as treatment/outcome networks \cite{hansen1982large}. 

However, finding suitable IVs is still difficult. Recently, \citet{yuan2022auto} proposed the Auto IV, which finds IVs $\hat{I}$ from candidates $X$ that satisfy Definition \ref{def_iv} by maximizing the mutual information (MI) between $\hat{I}$ and $T$ to ensure \textit{relevance}, and minimizing the conditional MI between $\hat{I}$, $Y$ given $T$ to ensure the \textit{restriction} criteria.

The advantages of IV-based methods are that \textbf{\textit{(i)}} no randomized data are required to address latent confounding, and \textbf{\textit{(ii)}} mature methods exist with good theoretical properties. However, it is difficult to find IV that satisfies the Definition \ref{def_iv}. In addition, if the IV is weak, i.e., has mild influences on the received treatments, the estimation will have a high variance even with large data.

\subsubsection{\textbf{Front-door Adjustment}} In addition, if the causal mechanism between the treatment $T$ and outcome $Y$ is known, i.e., all mediators $M$ are observable and unconfounded with  $T$ and $Y$, front-door adjustment can be used to address latent confounders \cite{pearl1995causal}. Specifically, based on the probability theory, we have
\begin{equation}
\label{eq:front}
    \mathbb{E}[Y|do(T)] = \mathbb{E}_{p_{(M|do(T))}}[Y|do(M)].
\end{equation}
Since no backdoor path exists between $T$ and $M$, $p(M|do(T)) = p(M|T)$. In addition, since $T$ blocks the backdoor path between $M$ and $Y$, $P(Y|do(M)) = \mathbb{E}_{P(T)}[P(Y|M,T)]$, where Eq. (\ref{eq:front}) is reduced to conditional relations measurable from the data. However, similar to IV-based methods, mediators that satisfy the front-door criterion are difficult to find. Therefore, \citet{xu2023causal} proposed to infer latent mediators that satisfy the front criterion from the covariate $X$ with the identifiable variational auto-encoder (iVAE) \cite{khemakhem2020variational}. 

\subsubsection{\textbf{Multiple Causes}} Finally, we consider the case of multiple treatments, where we are interested in estimating the combined causal effects of all the treatments in $T$ (e.g., prescribing bundled drugs) on $Y$. If we can determine that the latent confounders are shared among different treatments (i.e., single-cause ignorability \cite{wang2019blessings}), various methods can be used to address the confounding bias. The deconfounder-based methods prove that if latent variables $Z$ can be found that render different treatments conditional independent, controlling $Z$ adjusts for the confounding bias due to multi-cause confounders $C^{m}$ \cite{wang2019blessings}. The proof is simple and elegant: if $C^{m}$ are still active after conditioning on $Z$, they will render the treatments dependent (see Section \ref{sec:scm}), which results in contradiction. Linear models \cite{wang2020causal} and DNNs \cite{saini2019multiple,zhu2022deep,zou2020counterfactual} are used to estimate $Z$ from $T$. Recently, \citet{ma2021multi} proposed to learn latent $Z$ with latent clustering, which can well accommodate new treatments. Observing that under single-cause ignorability assumption, the data is unconfounded for every single cause, \citet{qian2021estimating} proposed to learn a single cause interventional model for each cause, and perturb the cause to generate counterfactually-augmented datasets, which they show are beneficial to learn multi-cause models.

\subsection{Inference-based Methods}

In this subsection, we introduce inference-based methods, which assume that even if confounders $C$ cannot be directly observed, we can observe their proxies $W$, which could be conducive to the inference of latent confounders to address confounding bias \cite{tchetgen2020introduction}. 

\subsubsection{\textbf{Proxy-based Methods}}
\label{sec:prox}
 Statistical methods generally assume simple forms of  $C$ and its causal relations with observed proxies $W$ \cite{pearl2012measurement}. However, even for the simplest relation, i.e., $C \rightarrow W$, directly controlling $W$ leaves the backdoor path $T \leftarrow C \rightarrow Y$ open, which cannot adjust for all the confounding bias. To address this, \citet{kuroki2014measurement} assumed that $W$ contains two independent views of $C$ to recover $p(W|C)$, such that $p(Y|do(T))$ can be identified from $p(W|C)$ and other observable relations. \citet{miao2018identifying}  relax the assumption, allowing for an unbiased estimation without recovering the confounder measurement error mechanism $p(W|C)$.

However, CI usually faces high-dimensional confounders and proxies $W$, where the statistical methods may fail to scale up to. Observing that even if $C$ is complex, only a small part of the information in $C$ is necessary to adjust for confounding (e.g., if $f(C)$ preserves the propensity score, i.e., $\mathbb{E}[T|C] = \mathbb{E}[T|f(C)]$, adjusting for $f(C)$ in Eq. (\ref{eq:back}) still gives an unbiased estimate of ATE/CATE \cite{rosenbaum1983central}), \citet{kallus2018causal} proposed to use matrix factorization (MF) to obtain low-rank components of $W$, which they show are better approximations of true confounders $C$. \citet{louizos2017causal} proposed the causal effect variational auto-encoder (CEVAE), which uses the VAE \cite{kingma2014auto} to recover the joint distribution $p(C,T,W,Y)$ and infer latent confounders $C$ from the observations $\{T, W, Y\}$. 

In addition to unbiasedness, other aspects need to be carefully considered as well. To address high variance due to non-overlapping covariate $W$ (i.e., certain values of $W$ appear only when $T=0$ or $T=1$, which is common if $W$ is high-dimensional), \citet{wu2021beta} proposed to map $W$ to low dimensional space with better overlapping w.r.t. the prognostic score \cite{hansen2008prognostic} based on the iVAE \cite{khemakhem2020variational}, which is sufficient for the identification of ATE/CATE. Furthermore, in certain cases, we cannot identify latent variables $Z$ that lead to an unbiased estimation of ATE/CATE. Therefore, \citet{hu2021generative} proposed an adversarial learning \cite{goodfellow2014generative}-based method to bound the error by finding the max/min values of the possible ATE via a generator and ensures that the distribution parameterized by the generator is faithful to the observations via another discriminator.

\subsubsection{\textbf{Covariate Disentanglement}}

If the proxy $W$ scrambles variables other than latent confounders $C$, various issues could be incurred if naively using the above-introduced proxy-based methods. To address this issue, covariate disentanglement (CD) methods are proposed to further scrutinize the latent variables $Z$ that generate the proxy $W$. Most methods assume that $W$ is generated from three types of latent variables: IVs $I$ (see Definition \ref{def_iv}), confounders $C$, and adjusters $A$, i.e., variables that causally influence only the outcome $Y$. Previous work has proven that controlling $C$ eliminates confounding bias, controlling $A$ could reduce estimation variance, while controlling IVs $I$ could \textit{increase} the variance \cite{myers2011effects,hu2023dbrnet}.

To address the issue, most methods rely on the statistical property between $\{I,C,A\}$ and $\{T, Y\}$: IVs $I$ are correlated with only the treatment $T$, adjusters $A$ are correlated only with the outcome $Y$, while confounders $C$ are correlated with \textbf{both} $T$ and $Y$. To leverage this property, DR-CFR \cite{hassanpour2019learning} designs three encoders to infer three sets of latent variables $\hat{I}$, $\hat{C}$, $\hat{A}$ from $W$, which are learned by making $\hat{I}, \hat{C}$ predictive for $T$ (i.e., maximizing $p(T|\hat{I}, \hat{C})$) and $\hat{C}$, $\hat{A}$ predictive for $Y$ (i.e., maximizing $p(Y|\hat{A}, \hat{C})$). Similarly, TEDVAE \cite{zhang2021treatment} splits the encoder in CEVAE into three parts for $\hat{I},\hat{C},\hat{A}$, respectively, and maximizes $p(T|\hat{I}, \hat{C})$, $p(Y|T, \hat{A}, \hat{C})$ to achieve disentanglement. 

The disentanglement can also be achieved by relying on other properties. For example, assuming $I,C,A$ are separable in the observed proxy $W$, AFS \cite{wang2023treatment} shows that $Z = \{C, A\}$, provided that the efficient influence curve of $Z$, $D^{eff}(Z)$ \cite{van2011targeted} is minimized. The empirical estimate of $D^{eff}(Z)$ is then used as the reward to learn a mask to select $C, A$ from $W$. In addition,  NICE \cite{shi2021invariant} uses invariant risk minimization (IRM) \cite{arjovsky2019invariant} to find all causal parents of $Y$ (including $C$, $A$), which can effectively exclude IVs from the control set. 

However, the above methods fail when latent post-treatment variables $M'$ are scrambled in the proxy $W$, as similar to $C$, post-treatment variables $M'$ can be correlated with \textbf{both} the treatment $T$ and the outcome $Y$, and can be the causal parents of $Y$. Recently, CiVAE \cite{zhu2024treatment} was proposed to disentangle $C$ from $M'$. Specifically, after individually identifying latent variables $\hat{Z}$ that generate $W$, independence tests are conducted for each pair of $\hat{Z}_{i}, \hat{Z}_{j}$, and the pairs with increased correlation after conditioning on $T$ are selected as confounders. In contrast, if both $\hat{Z}_{i}, \hat{Z}_{j}$ are post-treatment variables or one is a post-treatment variable and another is a confounder, the correlation will decrease after conditioning on $T$ (see \cite{pearl2009causality}).

\section{Causal Mediation Analysis}
\label{sec:med}

\begin{figure}[t]
\centering 
\includegraphics[width=0.40\textwidth]{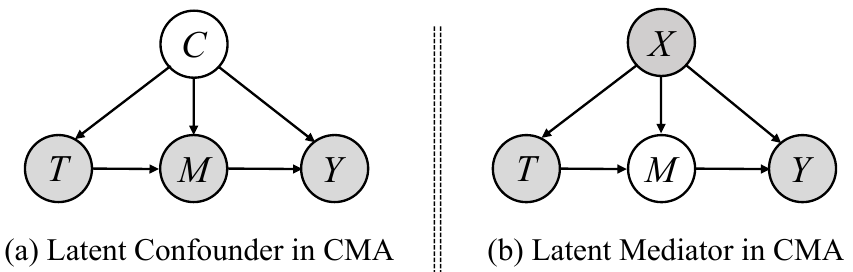}
\vspace{-2mm}
\caption{SCM for latent variables in CMA}
\vspace{-3mm}
\label{fig:med}
\end{figure}

\noindent Causal mediation analysis (CMA)~\cite{preacher2015advances} aims to understand the fine-grained causal mechanism between treatment $T$ and outcome $Y$ by identifying the effect mediated by another factor $M$, which mediates the causal effect from the treatment to the outcome.

\subsection{Brief Review of Traditional Methods}

Traditional CMA identifies the causal mediation effect based on the assumptions of \textit{measurable} confounders and mediators. Specifically, unobserved confounders are assumed away via the Sequential Ignorability assumption defined as follows:
\begin{assumption}
\label{ass:seq_ig}
\textbf{(Sequential Ignorability, SI).} \textbf{\textit{(i)}} $M(t), Y(t, m) \\ \indep T | X$; \textbf{\textit{(ii)}} $Y(t, m) \indep M(t) | T, X$ for all t, m.
\end{assumption}
\noindent Intuitively, SI assumption states no unobserved confounding between \textbf{\textit{(i)}} the treatment $T$ and the mediator $M$, \textbf{\textit{(ii)}} the mediator $M$ and the outcome $Y$, and \textbf{\textit{(iii)}} the treatment $T$ and the outcome $Y$. In addition, since $M$ is the factor of which the mediated effect is interested in, it should be observed and measurable.
\begin{assumption}
\label{ass:meas_med}
\textbf{(Measurable Mediator).} $M$ is observed.
\end{assumption}

\noindent With Assumptions \ref{ass:seq_ig}, \ref{ass:meas_med}, the causal effect mediated by $M$ in the form of natural indirect effect ($NIE$, see Eq. (\ref{eq:nie})) can be calculated as
\begin{equation}
\begin{aligned}
\label{eq:nie_est}
NIE = \mathbb{E}_{p(X)}\big[&\mathbb{E}_{p(M|T=1,X)}\left[Y|T=0,X,M\right] -\\
&\mathbb{E}_{p(M|T=0,X)}\left[Y|T=0,X,M\right]\big],
\end{aligned}
\end{equation}
which holds $T=0$ fixed on the direct path $T \rightarrow Y$, and change $T$ from $0$ to $1$ on the indirect path $T \rightarrow M \rightarrow Y$. The natural direct effect (NDE) of $T$ on $Y$ can be calculated as $ATE-NIE$. In practice, the conditional distributions required by $NIE$ and $NDE$ can be estimated using various methods, such as linear regression \citep{baron1986moderator}, logistic regression \citep{mackinnon2007mediation}, or machine learning techniques \citep{imai2010general}, such as decision trees and deep neural networks \citep{imai2010general, farbmacher2022causal}.

\subsection{Latent Confounders in CMA}

If latent confounders $C$ exist and are not included in $X$, the sequential ignorability assumption breaks and traditional methods that rely on Eq. (\ref{eq:nie_est}) to estimate $NIE$ will give biased results. To address the issue, various proxy-based methods are proposed. Here, an exemplar work is causal mediation analysis variational auto-encoder (CMAVAE)~\cite{cheng2022causal}, which assumes that the latent confounder $C$ confounds all pair-wise relations among $T, Y, M$, of which a noisy proxy $W$ can be observed (definition of proxy see Section \ref{sec:prox}). Inspired by CEVAE \cite{louizos2017causal}, they prove that the $NIE$ can be identified by estimating the joint distribution $p(C, W, M, T, Y)$, which are parameterized with DNNs, where the posterior distribution $q(C|W, M, T, Y)$ is obtained via variational inference \cite{kingma2014auto}. Finally, they sample latent confounders $C$ from $q(C|W, M, T, Y)$ to unbiasedly estimate $NIE$.

\subsection{Latent Mediation Analysis}

In this part, we discuss CMA with latent mediators, where Assumption \ref{ass:meas_med} fails, and Eq. (\ref{eq:nie_est}) cannot be used directly for estimation.

\subsubsection{\textbf{Circumvention-based Methods}}
Circumvention-based methods are difficult in CMA with latent mediators. Nevertheless, Derkach at. al.~\cite{derkach2019high} proposed a method without any utilization of observable proxies of latent mediators. They made a strong assumption that the distribution of $Y$ belongs to an exponential family $f(Y; \xi, \phi_Y)=exp{[Y\xi_Y-b(\xi_Y)]/a(\phi_Y)+c(Y, \phi_Y)}$, where $\xi_Y$ and $\phi_Y$ are modeled as functions of the latent mediator. They observe that $NIE$ can be represented by the parameters of the distribution. Therefore, to estimate the $NIE$, it is sufficient to estimate the parameters of $f(Y; \xi, \phi_Y)$, which is solved via an expectation-maximization (EM) algorithm: In each iteration of the algorithm, the expected latent mediators are first calculated, then the distribution parameters are updated via likelihood maximization. 
\subsubsection{\textbf{Proxy-based Methods}}
If the mediator of interest $M$ is not directly observed, utilizing its observed proxies $W$ is an effective method for $NIE$ estimation. Kuroki et al. ~\cite{kuroki2007graphical, kuroki2014measurement} showed that the $NIE$ of treatment $T$ on an outcome $Y$ could be identified in linear models given  two independent proxies of an unobserved mediator. In addition, Albert et al.~\cite{albert2016causal} proposed a maximum likelihood-based approach to estimate causal mediation effects with a continuous latent mediator measured by multiple observed proxies. Their method is based on fitting a generalized structural equation model (GSEM) \cite{muthen1984general} using an approximate Monte Carlo EM algorithm. The fitted GSEM is then used to estimate natural direct and indirect effects~\cite{pearl2012mediation}. In addition to the latent mediator, this approach also accommodates mediator-outcome confounding and mixed continuous and categorical outcomes. However, it relies on parametric modeling assumptions and may be computationally intensive. Recently, Sun et al.~\cite{sun2021bayesian} proposed a joint modeling approach that incorporates multiple latent mediators and a survival outcome. Specifically, a Bayesian approach with a Markov chain Monte Carlo algorithm is developed to perform an efficient estimation of $NIE$. 

\begin{figure}[t]
\centering 
\includegraphics[width=0.40\textwidth]{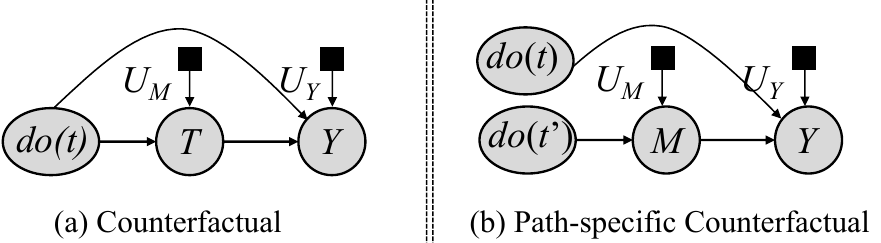}
\vspace{-2mm}
\caption{SCM for (path-specific) counterfactuals}
\vspace{-3mm}
\label{fig:ate12}
\end{figure}

\section{Counterfactual Analysis}
\label{sec:count}
In this section, we focus on counterfactual reasoning, which aims to explain the outcomes of a specific individual if a different treatment was taken in the past. According to SCM, exogenous variables $U$ contain individual varieties (see Section \ref{sec:scm}). Therefore, counterfactual reasoning under SCM is naturally a latent variable problem \cite{pearl2009causality}. Please note that throughout this section, we assume Sequential Ignorability (see Assumption \ref{ass:seq_ig}) holds, so that we can devote the main discussions to the latent exogenous variables.

\subsection{Overview}
With a pre-defined SCM $\mathbf{G}$, counterfactuals generally have the following form: $\mathbb{E}[Y_{U}(T=t')|W, T=t]$, where $W$ is the evidence (observed values for variables in $\mathbf{G}$) and $T=t$ is the observed treatment. Here, we use the subscript $U$ in $Y_{U}(T=t')$ to denote the dependence of the PO $Y(T)$ on exogenous variables $U$. The counterfactual inference involves three steps as follows:
\begin{itemize}[leftmargin=0.5cm]
    \item \textbf{\textit{(abduction)}} The prior of $U$, i.e., $p(U)$, is updated into posterior $p(U|W, T)$ based on the observed $W$ and $T=t$.
    \item \textbf{\textit{(action)}}  Structural equation $f_{T}$ is substituted with $f_{do}(T)=t^{\prime}$.
    \item \textbf{\textit{(prediction)}} The outcome $Y$ is computed with $p(U|W, T=t^{\prime})$, $f_{do}(T)=t^{\prime}$, and other structural equations in $\mathcal{F}$.
\end{itemize}
The key to counterfactual inference lies in the abduction of latent exogenous variables $U$, as the other two steps are straightforward.

\subsection{Circumvention-based Methods}

We can circumvent latent exogenous variables $U$ if the counterfactuals of interest are not required to be qualitatively determined. For example, when studying counterfactual fairness of ML models, we only need to judge whether two counterfactuals are the same:
\begin{equation}
\label{eq:cf}
\mathbb{E}[\hat{Y}_{U}(T=t')|W=w,T=t] \overset{?}{=}  \mathbb{E}[\hat{Y}_{U}(T=t)|W=w,T=t].
\end{equation}
Intuitively, Eq. (\ref{eq:cf}) asks that given evidence $W=w$ and $T=t$ (where $T$ could be the sensitive features such as race, gender, etc.), whether the prediction $\hat{Y}$ would be the same for a unit $U$ if $T$ is set to another value $t'$. \citet{kusner2017counterfactual} showed that Eq. (\ref{eq:cf}) holds when predictor $\hat{Y}$ does not use any descendant of $T$, which precludes the dependence of $\hat{Y}$ on $T$. Afterward, \citet{chiappa2019path} proposed path-specific counterfactual fairness (PSCF), which allows the causal influence of $T$ on $\hat{Y}$ along certain causal paths. For example, in the single mediator case, we may allow $T \rightarrow M \rightarrow \hat{Y}$ while forbid $T \rightarrow \hat{Y}$, where $M$ is called a resolving variable \cite{kilbertus2017avoiding}. In this case, the counterfactual question can be formulated via NPOs as follows:
\begin{equation}
\label{eq:pscf}
 \mathbb{E}[\hat{Y}_{U}(M(t),T=t')|W=w,T=t] \overset{?}{=}  \mathbb{E}[\hat{Y}_{U}(M(t), T=t)|W=w,T=t]. 
\end{equation}
If Eq. (\ref{eq:pscf}) holds, the predictor $\hat{Y}$ is precluded from using the descendants of $T$ along the unfair paths ($T$ itself included) \cite{nabi2018fair}.

\subsection{Inference-based Methods}

In other cases, when counterfactuals need to be calculated or bounded, inference-based methods become more useful \cite{zhu2023path}. Observing that exogenous variables $U$ satisfy exactly the non-descendant requirement of $T$ while containing all individual information, \citet{kusner2017counterfactual} assumed linear structural equations and fitted linear additive models on the observed data, where the error terms are viewed as the estimand of $U$ and used for fair predictions. 
\citet{zuo2022counterfactual} further generalized \cite{kusner2017counterfactual} to the case of partially observed SCMs, under the assumption that $T$ has no endogenous ancestor. \citet{wu2019pc} proposed to bound Eq. (\ref{eq:pscf}) by dividing the $U$ space into equivalent regions via response functions \cite{balke1994counterfactual}, and search the upper and lower limit of Eq. (\ref{eq:pscf}) while making the response functions compatible with the observed $W$ and $T$. To achieve PSCF, \citet{chiappa2019path} proposed to use VAE to infer $U$ and use it to correct the  dataset by setting $T$ of all samples to the baseline value along the unfair paths.

\section{Causal Discovery}

\noindent Previous sections primarily focus on CI with a \textit{pre-defined} causal graph. However, when accurate causal relations cannot be obtained (e.g., lack of domain knowledge), it becomes imperative to automatically discover the causal relations from data via causal discovery (CD). In this section, we first introduce traditional CD methods, including constraint-based and score-based methods. We then discuss CD strategies when unobserved confounders exist.

\label{sec:discovery}

\subsection{Brief Review of Traditional Methods}

Causal discovery (CD) aims to infer causal relations among variables of interest $\mathcal{V}$ from the observational dataset, with the goal of constructing a causal graph $\mathbf{G}=(\mathcal{V}, \mathcal{E})$. Most traditional CD methods rely on the assumptions of faithfulness and causal sufficiency (which assume away unobserved confounders) as follows:

\begin{assumption}
\textbf{\textit{(Faithfulness)}.} If two disjoint sets of variables $\mathcal{M}$ and $\mathcal{N}$ are independent in the distribution $P$ when conditioning on $\mathcal{Z}$, then it implies that $\mathcal{M}$ and $\mathcal{N}$ are d-separated~\cite{spirtes2000causation} in the graph $\mathbf{G}$ conditioning on $\mathcal{Z}$, denoted as:
$\mathcal{M} \indep_{P}\mathcal{N}|\mathcal{Z}\Longrightarrow  \mathcal{M}	 \indep_{\mathbf{G}}\mathcal{N}|\mathcal{Z}$.
\end{assumption}

\begin{assumption}
\textbf{\textit{(Causal Sufficiency)}.}
For any two observed variables $V_i$ and $V_j$ in the data, all common causes must also be observed.
\end{assumption}

\noindent Generally, CD methods can be categorized into two classes: \textbf{\textit{(i)}} constraint-based and  \textbf{\textit{(ii)}} score-based methods~\cite{nogueira2022methods}. Constraint-based methods use conditional independence tests to identify edges in the graph based on the faithfulness assumption. For example, the Peter-Clark (PC) algorithm~\cite{spirtes2000causation} and its variants~\cite{wang2020towards,cui2016copula,le2016fast,biza2020tuning} first identify an undirected causal graph (i.e., \textit{skeleton}) by removing edges from a complete causal graph with conditional independence tests, and then determine the edge direction by a set of orientation propagation rules with V-structures and acyclicity property ~\cite{spirtes2000causation}. For example, consider a path in the skeleton $A - B - C$, where $A$ and $C$ are not adjacent. If $A$ and $C$ became dependent conditioning on $B$, then the PC algorithm orients the edges as $A \rightarrow B \leftarrow C$ based on the property of V-structures (see Section \ref{sec:scm}). In contrast, score-based algorithms~\cite{chickering2002optimal,ramsey2017million,wang2017permutation} aim to identify the best candidate graph by maximizing a fitness score, such as the Bayesian Information Criterion (BIC), to discover the causal graph from the data. 

\subsection{Proxy-based Methods}

There are a few proxy-based methods for CD with unobserved confounders. \citet{liu2023causal} studied causal discovery between two variables $T$, $Y$ with a latent confounder $U$. Assuming a proxy $W$, i.e., a causal descendant of $U$, can be observed, they discretize $W$ and use \cite{miao2018identifying} introduced in Section \ref{sec:prox} to estimate the causal effect and judge whether an edge exists between $T$ and $Y$. However, such proxies may not exist in reality. Recently, \cite{liu2024causal} introduced time series data to address the issue, where each variable is assumed to be its causal parent in the next time step,  serving as the proxy $W$.

\subsection{Circumvention-based Methods}

\subsubsection{\textbf{Constraint-based Methods}} If unobserved confounders exist, the causal sufficiency assumption will not hold, and naive independence tests (i.e., correlation) cannot indicate causal relations among variables of interest. To address this issue, \citet{spirtes2013causal} proposed the FCI algorithm, which extends the PC algorithm by introducing three more relations (in addition to $X \rightarrow Y$) to model the uncertainty regarding confounders: \textbf{\textit{(i)}} $X \leftrightarrow Y$ indicates the presence of unmeasured confounders; \textbf{\textit{(ii)}} $X \circ \rightarrow Y$ represents that either $X$ causes $Y$ or there are unmeasured confounders; \textbf{\textit{(iii)}} $X  \circ - \circ Y$ can represent any of the following scenarios: (1) $X$ causes $Y$, (2) $Y$ causes $X$, or (3) there are unmeasured confounders, where a new orientation rule \cite{spirtes2000causation,zanga2022survey} is used to orient edges. 

Subsequent research has been proposed to extend FCI from various perspectives, such as with enhanced efficiency~\cite{colombo2012learning,spirtes2001anytime,colombo2014order}, tailored for sparse causal graphs~\cite{strobl2018fast}, improved scalability~\cite{raghu2018comparison}, and incorporating different conditional independence tests~\cite{jabbari2017discovery}.

\vspace{-2mm}
\subsubsection{\textbf{Score-based Methods}} Score-based algorithms, e.g., GES and Fast GES (FGES)~\cite{colombo2012learning}, find optimal causal graph by greedily adding and deleting edges based on predefined scores measuring the fitness of a graph on observational data. However, they face challenges when latent confounders exist (i.e., causal insufficiency). To address the issue, the recent trend is to use confounders-robust constraint-based methods such as FCI to correct the bias. However, it underperforms when the sample sizes are small due to an inaccurate estimation of the independence relations \cite{spirtes2013causal}. The Greedy FCI (GFCI) algorithm~\cite{ogarrio2016hybrid} combines the strengths of both approaches. It uses GES to identify a supergraph of the skeleton, then employs FCI to prune the supergraph and determine the orientations to handle unmeasured confounders. This integration enhances performance while maintaining asymptotic correctness under causal insufficiency. However, GFCI's scoring function cannot be applied to mixed variables, which is addressed by the Bayesian Constraint-Based Causal Discovery (BCCD) algorithm~\cite{claassen2012bayesian} via utilizing a hybrid constraint and score-based approach for causal search.

\section{Future Directions}
\label{sec:future}

In this section, we discuss several promising directions to further advance causal inference studies with latent variables and discuss new opportunities in the era of large language models (LLM).

\subsection{On Theories and Model Design}

Firstly, there has been a growing interest in causal representation learning \cite{scholkopf2021toward}, which aims to develop models capable of automatically extracting and representing causal concepts and relations from data. An in-depth study of causal representation learning with latent variables would be interesting in real-world applications. 

Additionally, integrating multi-modal information of the unit, such as textual \cite{veitch2020adapting, zhu2022mutually}, visual \cite{yang2021causalvae}, and sensor data \cite{tsapeli2015investigating}, offers opportunities to compensate for the absence of observation in a single modality, which also increases the chance of finding applicable circumvention or inference methods to address the latent variable. 


Furthermore, improving the interpretation \cite{du2019techniques, chu2024task, guan2024ufid} (especially towards latent variables) is essential to foster trust and transparency in causal learning systems with latent variables, allowing users to comprehend and validate causal conclusions effectively. 

Finally, exploring uncertainty quantification techniques \cite{abdar2021review}, such as conformal prediction \cite{shafer2008tutorial}, can provide valuable information on the reliability and robustness of CI under latent variables, facilitate more informed decision-making, and provide  pessimistic/optimistic bounds when exact causal effects cannot be identified. 

\subsection{Opportunities in the LLM Era}

\noindent Recently, large language models (LLM) exhibit remarkable in-context learning and reasoning capabilities  \cite{wan2024bridging,wang2023knowledge,ren2024survey,zhu2023collaborative,zhao2023survey}. Although LLM itself is nowhere causal \cite{zevcevic2023causal} (after all, it still fits conditional distributions parameterized by transformer networks on corpora \cite{vaswani2017attention}), recent research has shown some promising results of LLMs to facilitate CI, e.g., causal reasoning \cite{jin2023cladder}, counterfactual analysis \cite{zhang2023if}, and causal discovery \cite{ban2023query,cohrs2023large,kiciman2023causal}. For example, \citet{jin2023cladder} showed that when provided with few-shot examples with chain-of-thought (CoT) \cite{wei2022chain} causal reasoning steps in the prompts, LLMs can construct causal graphs, formulate causal questions with the two frameworks and manage to solve it with observational data. 

Based on the above examples, we speculate that LLMs can also provide opportunities to advance CI with latent variables. Here, we provide the following interesting future perspectives. \textit{\textbf{(i)}} First, it is promising to see LLM facilitate the automatic identification of important latent variables that could be neglected by human beings. As such, issues of neglecting important variables can be prevented in advance. \textit{\textbf{(ii)}} In addition, if the absence of important variables is inevitable, LLM may have the potential to reason with new strategies to circumvent or infer the variables from proxy based on the reasoning ability to the causal relation of latent variables and observed variables at hand (i.e., automatic causal discovery).
\textit{\textbf{(iii)}} Furthermore, LLM may provide a usable and user-friendly interpretation of latent variable models for CI \cite{wu2024usable}, as well as how biases are generated and eliminated. \textit{\textbf{(iv)}} Finally, recent advances in multi-modal LLM \cite{driess2023palm} are also promising to systematically consider multi-modal features of a unit, where the more comprehensive causal graph can be established by the LLM to increase the chance of finding good solutions to address the latent variables.

\section{Conclusions}
\label{sec:conclu}

\noindent In this survey, we review recent advances in causal inference (CI) with latent variables, covering four main CI tasks, i.e., causal effect estimation, causal mediation analysis, counterfactual reasoning, and causal discovery. We start by briefly reviewing CI methods where important variables are assumed to be observed. Then, under the new taxonomy of inference-based and circumvention-based methods, we introduce methods that account for the absence of crucial variables. Furthermore, we generalize the above method to graphs, an important area for machine learning. Finally, we discuss future perspectives, especially the new opportunity in the LLM era.

\section*{Acknowledgment}

This work was supported in part by the National Science Foundation (NSF) under grants IIS-2006844, IIS-2144209, IIS-2223769, IIS-2316306, CNS-2154962, and BCS-2228534; the Commonwealth Cyber Initiative Awards under grants VV-1Q23-007, HV-2Q23-003, and VV-1Q24-011; the JP Morgan Chase Faculty Research Award; the Cisco Faculty Research Award; and Snap gift funding. 

\newpage

\appendix

\noindent \textbf{\huge Appendix}

\section{Generalization to Graph Data}

Causal inference on graph data (e.g., social networks) naturally faces unique challenges compared with traditional tabular data due to the intrinsic interconnection and interactions among units under study. In the last few decades, there have been substantial efforts in marrying causal inference with graph mining \cite{ma2022learning_AI,ma2023causal,dai2023graph}, where latent variables still severely impede the robustness and trustworthiness of causal conclusions. Here, we extend the methodology introduced in the main paper to graph data.

\vspace{1mm}
\noindent\textbf{Treatment Effect Estimation.} Estimating treatment effect on graphs inevitably requires particular method to handle the challenges brought by the graph structure. Studies in this area mainly include the following branches: \textbf{\textit{(i)}} \textit{Proxies including graph structure}: although latent confounders on graphs are easily neglected by regular methods, fortunately, the graph structure itself can serve as proxies for the latent confounders in many cases \cite{guo2020learning,ma2021deconfounding,guo2020ignite}. \textbf{\textit{(ii)}} \textit{Circumvention-Based Methods}: Under certain circumstances, graph structure affects the treatment assignments and plays the role of an instrumental variable \cite{ma2023look}. Therefore, IV-based causal effect estimation approaches can be applied. \textbf{\textit{(iii)}} \textit{Interference}: one major issue of treatment effect estimation on graphs is that there often exists interference between connected units (graph nodes), i.e., the treatment of one unit may causally influence the outcome of other units. This, however, violates the SUTVA assumption \cite{pearl2009causality} in traditional causal inference. There have been numerous explorations \cite{ma2021causal,ma2022learning_hyper,fatemi2020minimizing,huang2023modeling} in this problem, covering different types of graphs.

\vspace{1mm}
\noindent\textbf{Counterfactual Analysis.} On graphs, counterfactual reasoning targets on generating a different graph under certain circumstances different from the factual one. As the graph structure is involved, counterfactual analysis on graphs often involves additional considerations regarding the causal relations among nodes, as well as the discrete and unorganized structural space. Various investigations have been conducted for this problem, including different goals such as generalization \cite{sui2022causal,li2022out}, explanation \cite{tan2022learning,prado2023survey,ma2022clear,guan2023xgbd,jaimini2022causalkg}, and fairness \cite{ma2022learningfair,dong2022fairness,guo2023towards} in many important applications. 

\noindent\textbf{Causal Discovery.} The nature of graphs makes them closely associated with causal relations. Related causal discovery work in this area mainly includes \textbf{\textit{(i)}} methods based on classical graphical models \cite{glymour2019review}, which rely on causal graphical models and have been the mainstream of causal discovery; \textbf{\textit{(ii)}} methods based on learnable graph adjacency matrices in neural networks \cite{zheng2018dags,yang2021causalvae}, which discover the causal relations inside data by learning an $N\times N$ adjacency matrix for a causal graph with $N$ variables; \textbf{\textit{(iii)}} methods based on graph neural networks (GNNs) \cite{wang2023hierarchical,yu2019dag,ng2019graph}, which explicitly leverage GNN techniques to facilitate causal discovery.

\balance
\bibliographystyle{ACM-Reference-Format}
\bibliography{survey}

\end{document}